\documentclass[10pt,twocolumn,letterpaper]{article}

\usepackage{cvpr}
\usepackage{times}
\usepackage{epsfig}
\usepackage{graphicx}
\usepackage{amsmath}
\usepackage{amssymb}

\usepackage{color, soul}
\usepackage{bbm}
\usepackage{amsmath,amssymb}
\usepackage{bm}
\usepackage{caption}
\usepackage{dsfont}

\DeclareMathOperator*{\argmin}{\arg\!\min}
\usepackage{hhline}
\usepackage{subcaption}
\usepackage{mathtools}
\usepackage{multirow}
\usepackage{hhline}
\usepackage{enumitem}

\usepackage[pagebackref=true,breaklinks=true,letterpaper=true,colorlinks,bookmarks=false]{hyperref}

 \cvprfinalcopy 

\def\cvprPaperID{2332} 

\ifcvprfinal\fi
\begin{document}

\title{CERN: Confidence-Energy Recurrent Network for Group Activity Recognition}

\author{Tianmin Shu$^1$, Sinisa Todorovic$^2$ and Song-Chun Zhu$^1$\\
$^1$ University of California, Los Angeles $^2$Oregon State University\\
{\tt\small tianmin.shu@ucla.edu sinisa@onid.orst.edu sczhu@stat.ucla.edu}
}

\def \rb {\bm{r}} 
\def \sb {\bm{s}} 
\def \cb {\bm{c}} 
\def \hb {\bm{h}} 
\def \wb {\bm{w}} 
\def \bb {\bm{b}} 
\def \nb {\bm{n}} 

\def \xb {{\bf{x}}}
\def \ob {{\bf{w}}}
\def \pb {{\bf{p}}}
\def \lb {{\boldsymbol{\lambda}}}
\def \psib {{\boldsymbol{\psi}}}

\newtheorem{prop}{Proposition}

\DeclarePairedDelimiter{\ceil}{\lceil}{\rceil}

\maketitle

\begin{abstract}
This work is about recognizing human activities occurring in videos at distinct semantic levels, including individual actions, interactions, and group activities. The recognition is realized using a two-level hierarchy of Long Short-Term Memory (LSTM) networks, forming a feed-forward deep architecture, which can be trained end-to-end. In comparison with existing architectures of LSTMs, we make two key contributions giving the name to our approach as Confidence-Energy Recurrent Network -- CERN. First, instead of using the common softmax layer for prediction, we specify a novel energy layer (EL) for estimating the energy of our predictions. Second, rather than finding the common minimum-energy class assignment, which may be numerically unstable under uncertainty, we specify that the EL additionally computes the p-values of the solutions, and in this way estimates the most confident energy minimum. The evaluation on the Collective Activity and Volleyball datasets demonstrates: (i) advantages of our two contributions relative to the common softmax and energy-minimization formulations and (ii) a superior performance relative to the state-of-the-art approaches. 
\end{abstract}

 \begin{figure}[t!]
      \centering
      \includegraphics[width = 1.0\linewidth]{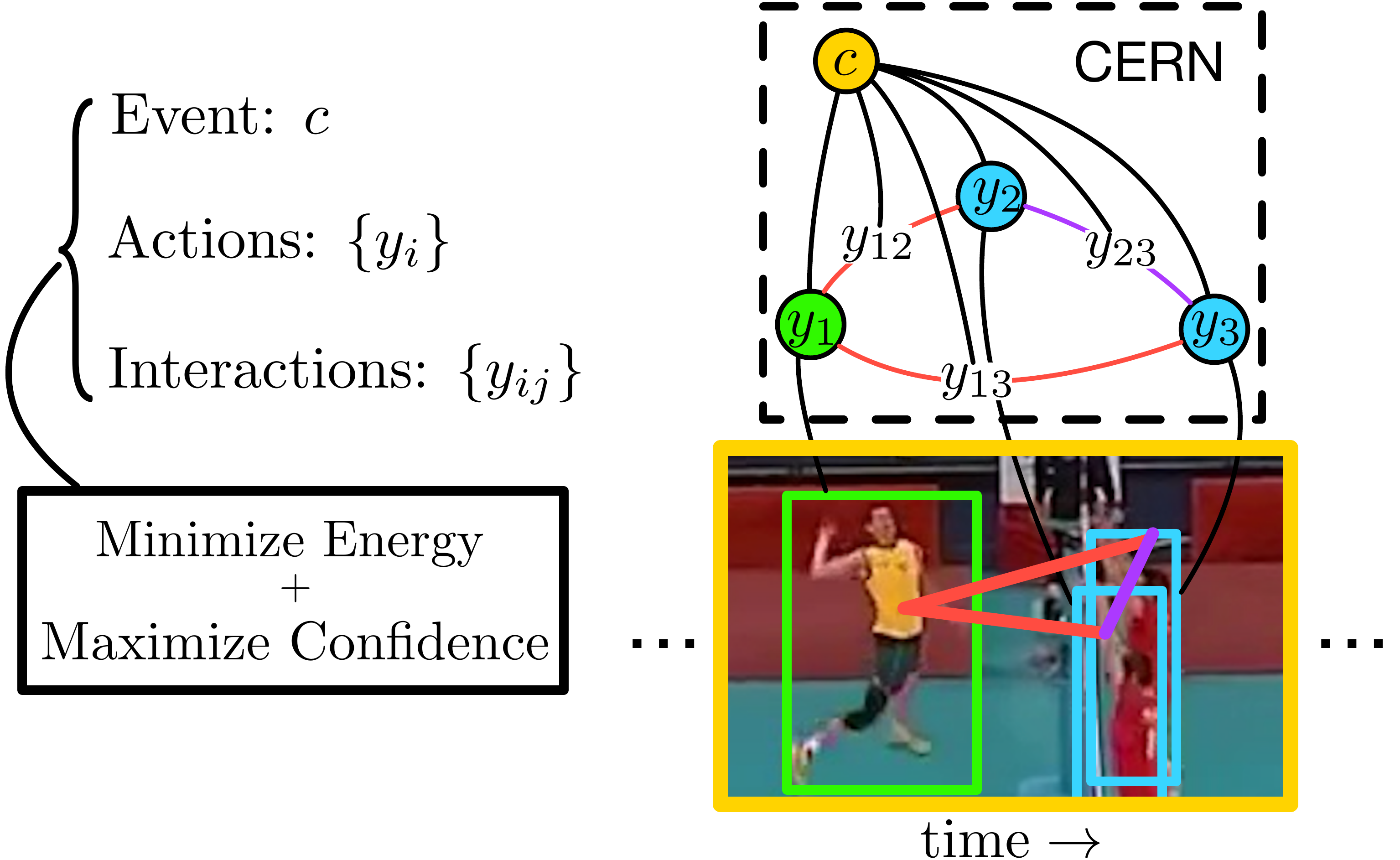}
      \caption{Our CERN represents a two-level hierarchy of LSTMs grounded onto human trajectories, where the LSTMs predict individual actions $\{y_i\}$, human interactions $\{y_{ij}\}$, or the event class $c$ in a given video. CERN outputs an optimal configuration of LSTM predictions which jointly minimizes the energy of the predictions and maximizes their confidence, for addressing the brittleness of cascaded predictions under uncertainty. This is realized by extending the two-level hierarchy with an additional energy layer, which can be trained in an end-to-end fashion.}
      \label{fig:intro}
   \end{figure}

\section{Introduction}
   
This paper addresses activity recognition in videos, each showing a group activity or event (e.g., spiking in volleyball) arising as a whole from a number of individual actions (e.g., jumping) and human interactions (e.g., passing the ball). Our goal is to recognize events, interactions, and individual actions, for settings where training examples of all these classes are annotated. When ground truth annotations of interactions are not provided in training data, we only pursue recognition of events and actions.

Recent deep architectures \cite{Ibrahim2016, Ramanathan2016}, representing a multi-level cascade of Long Short-Term Memory (LSTM) networks \cite{Hochreiter1997}, have shown great promise in recognizing video events. In these approaches, the LSTMs at the bottom layer are grounded onto individual human trajectories, initially obtained from tracking. These LSTMs are aimed at extracting deep visual representations and predicting individual actions of the respective human trajectories. Outputs of the bottom LSTMs are forwarded to a higher-level LSTM for predicting events. All predictions are made in a feed-forward way using the softmax layer at each LSTM. Such a hierarchy of LSTMs is trained end-to-end using backpropagation-through-time of the cross-entropy loss. 

Motivated by the success of these approaches, we start off with a similar two-level hierarchy of LSTMs for recognizing individual actions, interactions, and events. We extend this hierarchy for producing more reliable and accurate predictions in the face of the uncertainty of the visual input.   

Ideally, the aforementioned cascade should be learned to overcome uncertainty in a given domain (e.g., occlusion, dynamic background clutter). However, our empirical evaluation suggests that existing benchmark datasets (e.g., the Collective Activity dataset \cite{Choi2009} and the Volleyball dataset \cite{Ibrahim2016}) are relatively too small for a robust training of all LSTMs in the cascade. Hence, in cases that have not been seen in the training data, we observe that the feed-forwarding of predictions is typically too brittle, as errors made at the bottom level are directly propagated to the higher level. One way to address this challenge is to augment the training set. But it may not be practical as collecting and annotating group activities is usually difficult.

As shown in Fig.~\ref{fig:intro}, we take another two-pronged strategy toward more robust activity recognition that includes: 
\begin{enumerate}[itemsep=-5pt,topsep=2pt, partopsep=1pt]
\item Minimizing {\bf energy} of all our predictions at the different semantic levels considered, and
\item Maximizing {\bf confidence} (reliability) of the predictions. 
\end{enumerate}
Hence the name of our approach -- Confidence-Energy Recurrent Network (CERN).

Our first contribution is aimed at mitigating the brittleness of the direct cascading of predictions in previous work. We specify an energy function for capturing dependencies between all LSTM predictions within CERN, and in this way enable recognition by energy minimization. Specifically, we extend the aforementioned two-layer hierarchy of LSTMs with an additional energy layer (EL) for estimating the energy of our predictions. The EL replaces the common softmax layer at the output of LSTMs. Importantly, this extension allows for a robust, energy-based, and end-to-end training of the EL layer on top of all LSTMs in CERN.

\begin{figure}[t!]
      \centering
      \includegraphics[width = 0.9\linewidth]{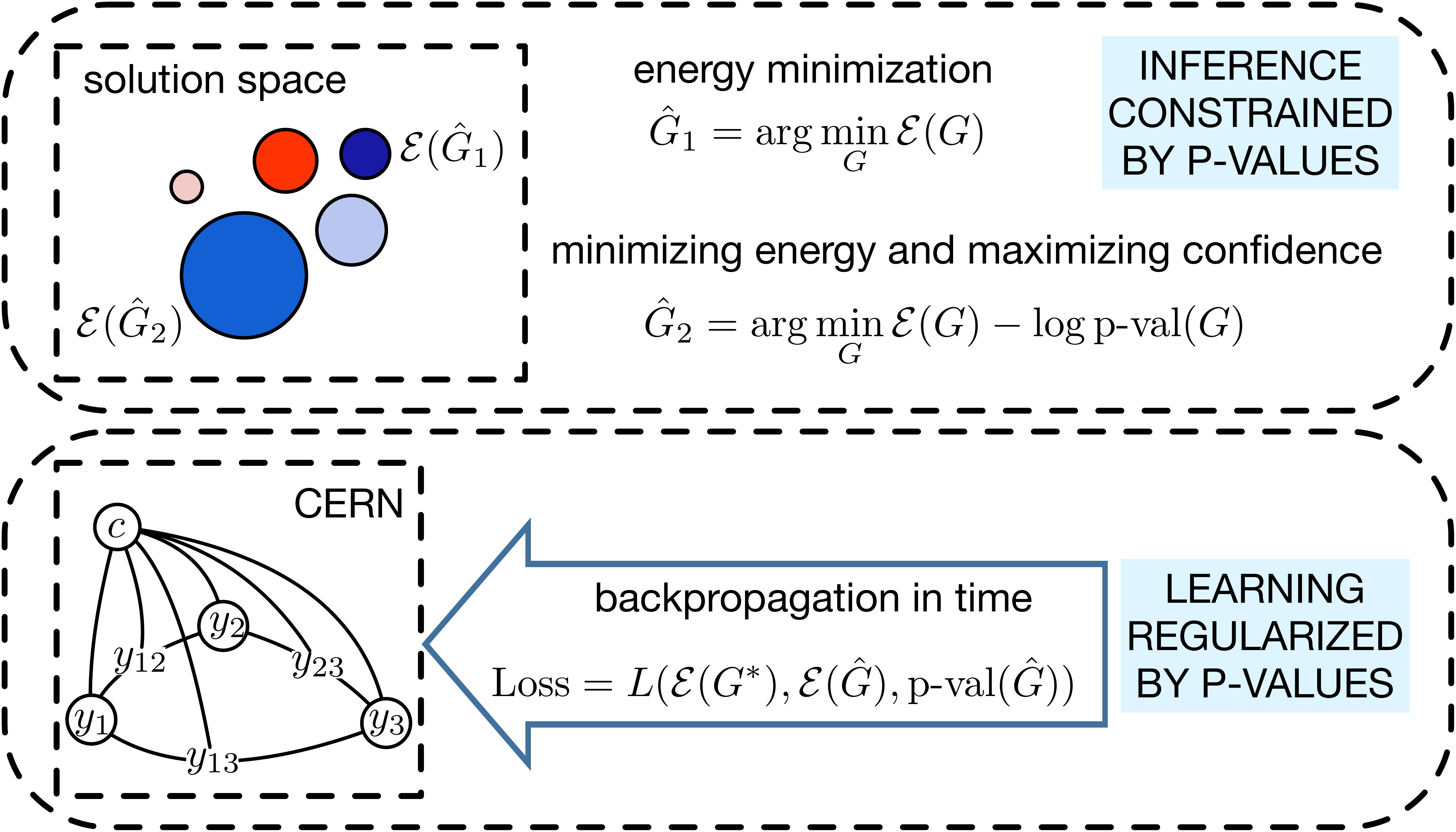}
      \caption{(top) An imaginary illustration of the solution space where each circle represents a candidate solution. The colors and sizes of the circles indicate the energy (red:high, blue:low) and confidence (the larger the radius the higher confidence) computed by the energy layer in CERN. A candidate solution $\hat{G}_1$ has the minimum energy, but seems numerically unstable for small perturbations in input. A joint maximization of confidence and minimization of energy gives a different, more confident solution $\hat{G}_2$. Confidence is specified in terms of p-values of the energy potentials. (bottom) We formulate an energy-based loss for end-to-end learning of CERN. The loss accounts for the energy and p-values.}
      \label{fig:framework}
   \end{figure}

Our second contribution is aimed at improving the numerical stability of CERN's predictions under perturbations in the input, and resolving ambiguous cases with multiple similar-valued local minima. Instead of directly minimizing the energy, we consider more reliable solutions, as illustrated in Fig.~\ref{fig:framework}. The reliability or confidence of solutions is formalized using the classical tool of a statistical hypothesis test \cite{Fisher1950} -- namely, p-values of the corresponding LSTM's hypotheses (i.e., class predictions). Thus, we seek more confident solutions by regularizing energy minimization with constraints on the p-values. This effectively amounts to a joint maximization of confidence and minimization of energy of CERN outputs. Therefore, we specify the EL to estimate the minimum energy with certain confidence constraints, rather than just the energy. 

We also use the energy regularized by p-values for robust deep learning. Specifically, we formulate an energy-based loss which not only accounts for the energy but also the p-values of CERN predictions on the training data. 

Our evaluation on the Collective Activity \cite{Choi2009} and Volleyball \cite{Ibrahim2016} datasets demonstrates: (i) advantages of the above contributions compared with the common softmax and energy-based formulations and (ii) a superior performance relative to the state-of-the-art methods.

In the following, Sec.~\ref{sec:prior_work} reviews prior work, Sec.~\ref{sec:CERNformulation} specifies CERN, Sec.~\ref{sec:energy}~and~\ref{sec:confidence} formulate the energy and confidence,  Sec.~\ref{sec:EL} describes the energy layer,  Sec.~\ref{sec:learning} specifies our  learning, and finally Sec.~\ref{sec:results} presents our results.

    \begin{figure*}
      \centering
      \begin{subfigure}{0.40\linewidth}
      \centering
      \includegraphics[width = 1.0\linewidth]{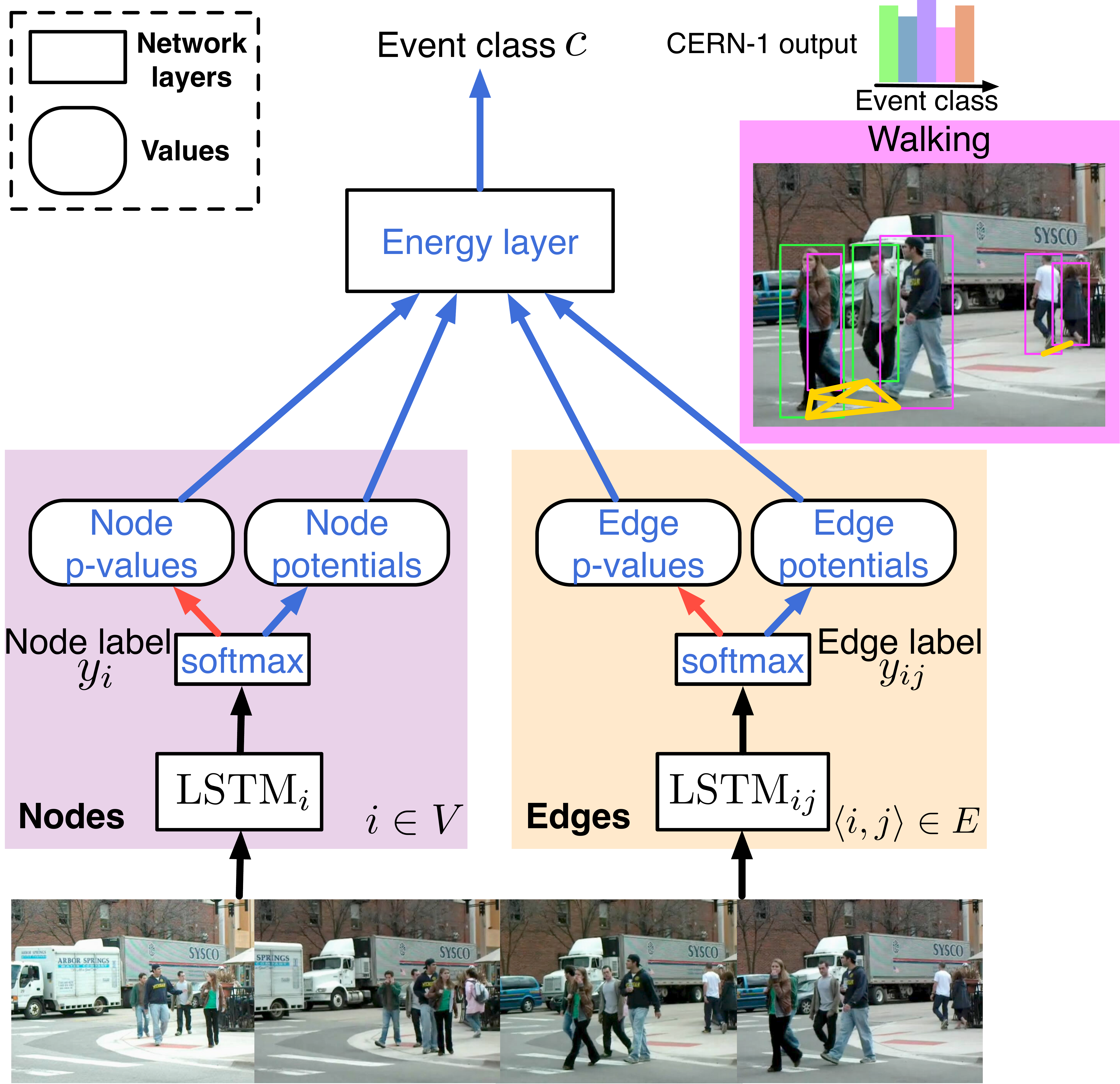}
      \caption{CERN-1}
      \label{fig:CERN-1}
      \end{subfigure}
      ~
      \begin{subfigure}{0.55\linewidth}
      \centering
      \includegraphics[width = 1.0\linewidth]{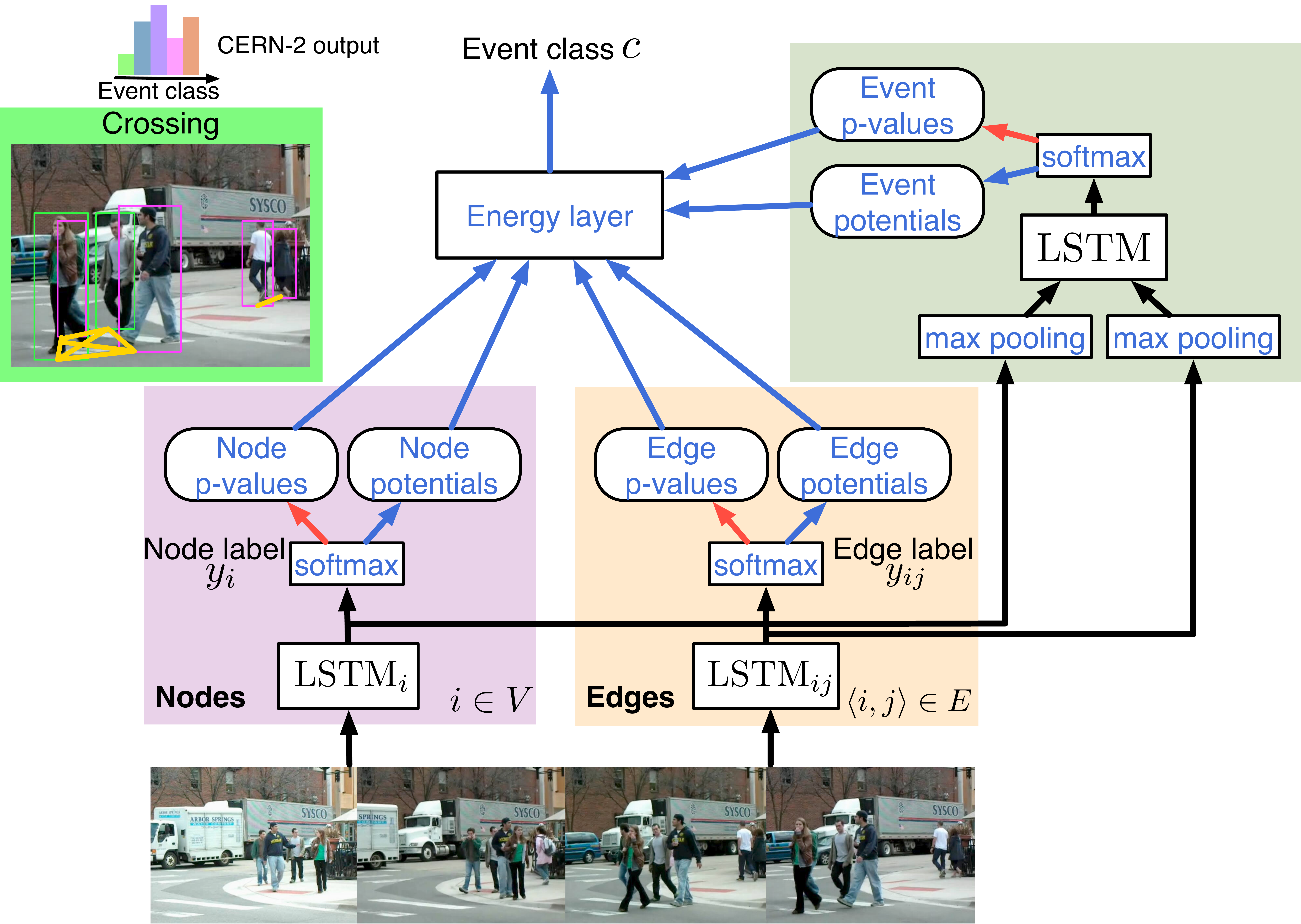}
      \caption{CERN-2}
       \label{fig:CERN-2}
      \end{subfigure}
      \caption{We specify and evaluate two versions of CERN. CERN is a deep architecture of LSTMs, which are grounded via CNNs to video frames at the bottom. The LSTMs forward their class predictions to the energy layer (EL) at the top. CERN-1 has LSTMs only at the bottom level which compute distributions of individual action classes (colored boxes) or distributions of interaction classes (colored links between green boxes). CERN-2 has an additional LSTM for computing the distribution of event (or group activity) classes. The EL takes the LSTM outputs, and infers an energy minimum with the maximum confidence. The figure shows that CERN-1 and CERN-2 give different results for the group activity \textit{crossing}. CERN-1 wrongly predicts \textit{walking}. CERN-2 typically yields better results for group activities that can not be defined only by individual actions.}
       \label{fig:architecture}
   \end{figure*}

\section{Related Work}\label{sec:prior_work}

{\bf Group activity recognition}. Group activity recognition often requires the explicit representation of spatiotemporal structures of group activities defined in terms of individual actions and pairwise interactions. Previous work typically used graphical models \cite{Lan2012PAMI,Lan2012,Ramananthan2013,Amer2014,Choi2014} or AND-OR grammar models \cite{Amer2012,Shu2015} to learn the structures grounded on hand-crafted features. Recent methods learn a graphical model, typically MRF \cite{Chen2015,Wu2016} or CRF \cite{Zhang2015,Jain2016,Liang2016}, using recurrent neural networks (RNNs). Also, work on group activity recognition \cite{Ibrahim2016,Deng2016} has demonstrated many advantages of using deep architectures of RNNs over the mentioned non-deep approaches.  Our approach extends this work by replacing the RNN's softmax layer with a new energy layer, and by specifying a new energy-based model that takes into account p-values of the network's predictions.

{\bf Energy-based learning}.  While energy-based formulations of inference and learning are common in non-deep group activity recognition \cite{Ramananthan2013,Amer2014,Choi2014,Shu2015}, they are seldom used for deep architectures. Recently, a few approaches have tried to learn an energy-based model \cite{LeCun2005,LeCun2006} using deep neural networks \cite{Belanger2016,Zhao2016}. They have demonstrated that energy-based objectives have great potential in improving the performance of structured predictions, especially when training data are limited. Our approach extends this work by regularizing the energy-based objective such that it additionally accounts for the confidence of predictions.

{\bf Reliability of Recognition}. Most energy-based models in computer vision have only focused on the energy minimization for various recognition problems. Our approach additionally estimates and regularizes inference with p-values. The p-values are specified within the framework of conformal prediction \cite{Shafer2008}. This allows the selection of more reliable and numerically stable predictions.

\section{Components of the CERN Architecture}\label{sec:CERNformulation}

For recognizing events, interactions, and individual actions, we use a deep architecture of LSTMs, called CERN, shown in Fig.~\ref{fig:architecture}. CERN is similar to the deep networks presented in \cite{Ibrahim2016,Jain2016}, and can be viewed as a graph $G = \langle V, E, c, Y\rangle$, where $V = \{i\}$ is the set of nodes corresponding to individual human trajectories, and $E = \{(i,j)\}$ is the set of edges corresponding to pairs of human trajectories. These human trajectories are extracted using an off-the-shelf tracker \cite{Danelljan2014}. Also, $c \in\{1,\cdots,C\}$ denotes an event class (or group activity), and $Y = Y^V \cup Y^E$ is the union set of individual action classes $Y^V=\{y_i:y_i \in \mathcal{Y}^V\}$ and human interaction classes $Y^E = \{y_{ij}:y_{ij} \in \mathcal{Y}^E\}$ associated with nodes and edges. 

In CERN, we assign an LSTM to every node and edge in $G$. All the node LSTMs share the same weights and all the edge LSTMs also have the same weights. These LSTMs use convolutional neural networks (CNNs) to compute deep features of the corresponding human trajectories, and output softmax distributions of individual action classes, $\psi^V(x_i, y_i)$, or softmax distributions of human interaction classes, $\psi^E(x_{ij}, y_{ij})$. The LSTM outputs are then forwarded to an energy layer (EL) in CERN for computing the energy $\mathcal{E}(G)$. Finally, CERN outputs a structured prediction $\hat{G}$ whose energy has a high confidence:

\begin{equation} 
\hat{G} = \arg\min_G \mathcal{E}(G) - \log \text{p-val}(G).
\label{eq:inference}
\end{equation}

As shown in Fig.~\ref{fig:architecture}, we specify and evaluate two versions of CERN. CERN-1 uses LSTMs for predicting individual actions and interactions, whereas the event class is predicted by the EL as in (\ref{eq:inference}). CERN-2 has an additional event LSTM which takes features maxpooled from the outputs of the node and edge LSTMs, and then computes the distribution of event classes, $\psi(c)$. The EL in CERN-2 takes all three types of class distributions as input -- specifically,  $\{\psi^V(x_i,y_i)\}_{i \in V}$, $\{\psi^E(x_{ij}, y_{ij})\}_{(i,j)\in E}$, and $\psi(c)$ -- and predicts an optimal class assignment as in (\ref{eq:inference}).

In the following, we specify $\mathcal{E}(G)$ and $\text{p-val}(G)$.

\section{Formulation of Energy } \label{sec:energy}

For CERN-1, the energy of $G$ is defined as
\begin{equation}
\setlength{\arraycolsep}{3pt}
\begin{array}{llll}
 \mathcal{E}(G) &\propto& \displaystyle \sum_{i 
\in V}  w^V_{c,y_i}\psi^V(x_i,y_i)  & \text{node potential}\\
 &+& \displaystyle \sum_{(i,j) \in E} w^E_{c,y_{ij}} \psi^E(x_{ij},y_{ij})  & \text{edge potential},
\end{array}
\label{eq:energy0}
\end{equation}
where $w^V_{c,y_i}$ and $w^E_{c,y_{ij}}$ are parameters,  $\psi^V(x_i,y_i)$ denotes the softmax output of the corresponding node LSTM, and $\psi^E(x_{ij},y_{ij})$ denotes  the softmax output of the corresponding edge LSTM (see Sec.~\ref{sec:CERNformulation}), and $x_i$ and $x_{ij}$ denote
visual cues extracted from respective human trajectories by a CNN as in \cite{Deng2016,Ibrahim2016}.

For CERN-2, the energy in (\ref{eq:energy0}) is augmented by the softmax output of the event LSTM, i.e., 
\begin{equation}
\setlength{\arraycolsep}{3pt}
\begin{array}{llll}
 \mathcal{E}(G) &\propto& \displaystyle \sum_{i 
\in V}  w^V_{c,y_i}\psi^V(x_i,y_i)  & \text{node potential}\\
 &+& \displaystyle \sum_{(i,j) \in E} w^E_{c,y_{ij}} \ \psi^E(x_{ij},y_{ij})  & \text{edge potential}\\
&+& \displaystyle w_c\psi(x, c) & \text{event potential},
\end{array}
\label{eq:energy0_2}
\end{equation}
where $x$ in $\psi(x, c)$ is the visual representation of all actions and interactions maxpooled from the outputs of the node LSTMs and edge LSTMs.

      \begin{figure}[t!]
      \centering
      \includegraphics[clip,trim={0 0 0 0}, width = 0.5\linewidth]{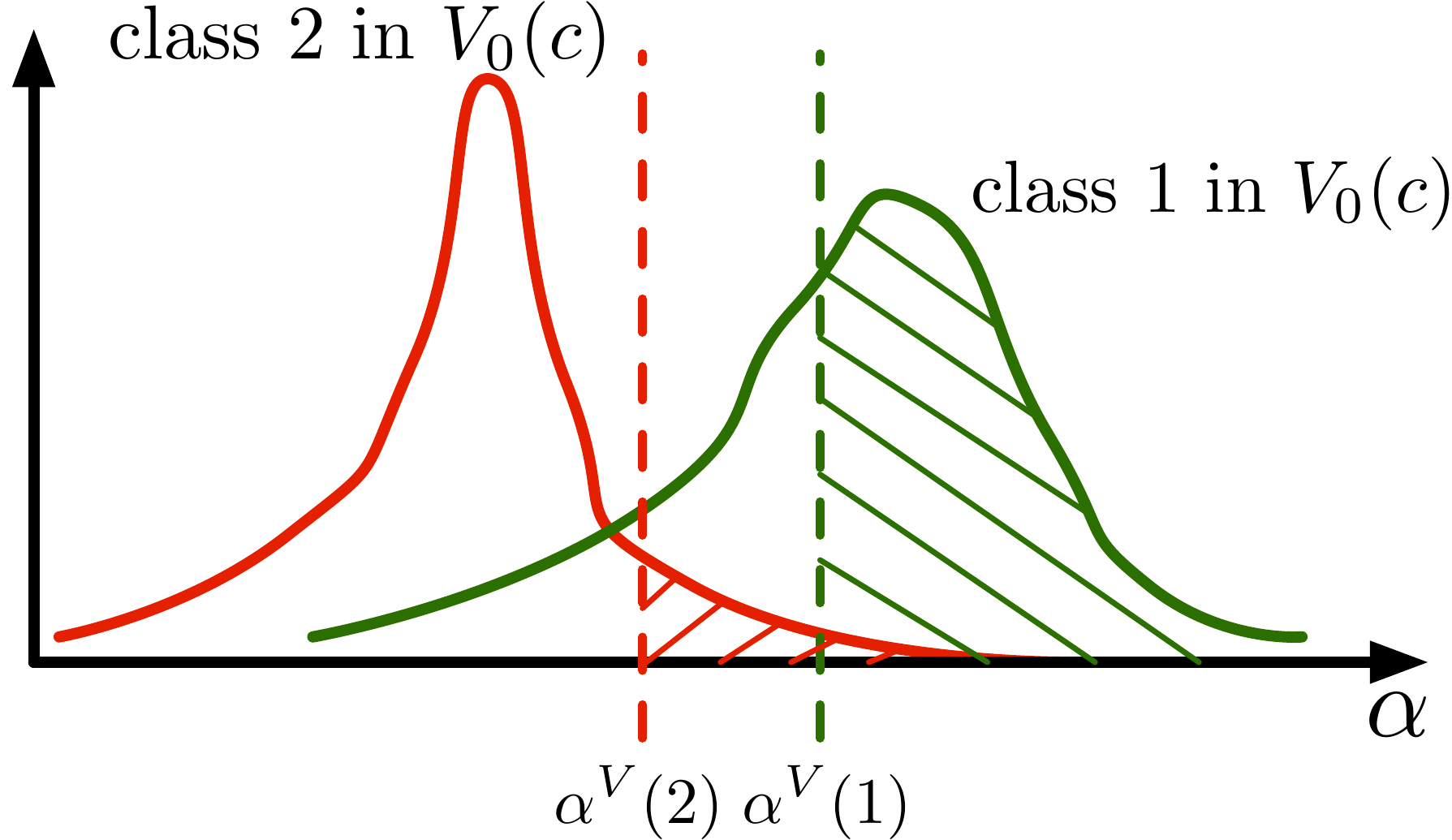}
      \vspace{-5pt}
      \caption{A simple illustration of the relationship between the nonconformity measure $\alpha$ of individual actions and the p-value, where the ratio of the dashed region to the whole area under the curve indicates the p-value. Clearly, for the given instance, action class 2 has a larger softmax output but action class 1 has a higher confidence. $V_0(c)$ is the training set of videos showing event $c$.}
      \vspace{-10pt}
      \label{fig:alpha_pvalue}
   \end{figure}

 \section{Formulation of Confidence } \label{sec:confidence}  
  
There are several well-studied ways to define the p-values \cite{Fisher1950}. In this paper, we follow the framework of conformal prediction \cite{Shafer2008}. Conformal prediction uses a nonconformity (dissimilarity) measure to estimate the extent to which a new prediction is different from the system's predictions made during training. Hence, it provides a formalism to estimate the confidence of new predictions based on the past experience on the training data. Below, we define the nonconformity measure, which is used to compute the p-values for LSTMs' predictions of individual actions, interactions, and events.

\subsection{Nonconformity Measure and P-values }

Given the node potential $\psi^V(x_i,y_i)$, we define a nonconformity measure for action predictions:
\begin{equation}
\alpha^V(y_i) =  1 - \frac{\psi^V(x_i,y_i)}{\sum_{y \in \mathcal{Y}^V} \psi^V(x_i,y)} = 1 - \psi^V(x_i,y_i),
\label{eq:alphan}
\end{equation}
where the above derivation step holds because $\psi^V(x_i,y_i)$ is the softmax output normalized over action classes. $\alpha^V(y_i) $ is used to estimate the p-value of predicting action class $y_i$ under the context of event class $c$ as
\begin{equation}
p_i^V(c, y_i) =  \frac{\sum_{{i^\prime} \in V_0(c)} \mathds{1}(y_{i^\prime} = y_i) \mathds{1}(\alpha^V(y_{i^\prime}) \geq \alpha^V(y_i))}{\sum_{{i^\prime} \in V_0(c)} \mathds{1}(y_{i^\prime} = y_i)}.
\label{eq:pval-node}
\end{equation}
where $\mathds{1}(\cdot)$ is the indicator, and $V_0(c)$ denotes the set of all human trajectories in training videos with ground truth labels $y_{i^\prime}$ and belonging to the ground truth event class $c$. From (\ref{eq:pval-node}), the LSTM prediction $\psi^V(x_i,y_i)$ is reliable -- i.e., has a high p-value -- when many training examples $i^\prime$ of the same class have larger nonconformity measures.

To better understand the relationship between the nonconformity measure and the p-value, let us consider a simple case illustrated in Fig.~\ref{fig:alpha_pvalue}. The figure plots the two distributions of nonconformity measures of two action classes in the training examples (green: class 1, red: class 2). Suppose that we observe a new instance whose softmax output indicates that action class 2 has a higher probability to be the true label, i.e., $\psi^V(x_i, 1) < \psi^V(x_i, 2)$, and $\alpha^V(1) > \alpha^V(2)$. From the two curves, however, we see that this softmax output is very likely to be wrong. This is because from Fig.~\ref{fig:alpha_pvalue} we have the p-values $p_i^V(c,1) > p_i^V(c,2)$, since a majority of training examples with the class 1 label have larger nonconformity measures than $\alpha^V(1)$, and hence class 1 is a more confident solution.

Similarly, given the softmax output of the edge LSTM, $\psi^E(x_{ij},y_{ij})$,  we specify a  nonconformity measure of predicting interaction classes:
\begin{equation}
\alpha^E_{ij}(y_{ij}) = 1 - \frac{\psi^E(x_{ij},y_{ij})}{\sum_{y \in \mathcal{Y}^E} \psi^E(x_{ij},y)}=1-\psi^E(x_{ij},y_{ij}),
\label{eq:alphae}
\end{equation}
which is then used to estimate the p-value of predicting interaction class $y_{ij}$ under the context of event class $c$ as
\begin{equation}
\begin{array}{l}
p^E_{ij}(c,y_{ij})\\
=  \displaystyle \frac{\sum_{(i^\prime, j^\prime) \in E_0(c)} \mathds{1}(y_{i^\prime j^\prime} = y_{ij})\mathds{1}(\alpha^E_{i^\prime j^\prime}(y_{i^\prime j^\prime}) \geq \alpha^E_{ij}(y_{ij}))}{\sum_{(i^\prime, j^\prime) \in E_0(c)}   \mathds{1}(y_{i^\prime j^\prime} = y_{ij})},
\end{array}
\label{eq:pval-edge}
\end{equation}
where $E_0(c)$ denotes the set of all pairs of human trajectories in training videos with ground truth labels $y_{i^\prime j^\prime}$ and belonging to the ground truth event class $c$. From (\ref{eq:pval-edge}), the LSTM prediction $\psi^E(x_{ij},y_{ij})$ has a high p-value when many training examples $(i^\prime, j^\prime)$ in $E_0(c)$ have larger nonconformity measures.

Finally, in CERN-2, we also have the LSTM softmax output $\psi(x,c)$, which is used to define a nonconformity measure for event predictions:
\begin{equation}
\alpha(c) =  1 - \frac{\psi(x,c)}{\sum_{c \in \mathcal{C}} \psi(x,c)}=1-\psi(x,c),
\label{eq:alphan}
\end{equation}
and the p-value of predicting event class $c$ as
\begin{equation}
p(c) =  \frac{\sum_{v \in V_0} \mathds{1}(c_v = c) \mathds{1}(\alpha(c_v) \geq \alpha(c))}{\sum_{v \in V_0} \mathds{1}(c_v = c)}.
\label{eq:pval-event}
\end{equation}
where $V_0$ denotes the set of all training videos.

\subsection{Confidence of the Structured Prediction $G$}

To define the statistical significance of the hypothesis $G$  among other hypotheses (i.e., possible solutions), we need to combine the p-values of predictions assigned to nodes, edges and the event of $G$. More rigorously, for specifying the p-value of a compound statistical test, $\text{p-val}(G)$, consisting of multiple hypotheses, we follow the Fisher's combined hypothesis test \cite{Fisher1950}. The Fisher's theory states that $N$ independent hypothesis tests, whose p-values are $p_1, \cdots p_N$, can be characterized by a test statistic $\chi^2_{2N}$ as
\begin{equation}
\chi^2_{2N} = -2\sum_{n=1}^N\log p_n,
\label{eq:Fisher}
\end{equation}
where the statistic $\chi^2_{2N}$ is proved to follow the $\chi^2$ probability distribution with $2N$ degrees of freedom. From (\ref{eq:Fisher}), it follows that minimization of the statistic $\chi^2_{2N}$ will yield the maximum p-value characterizing the  Fisher's combined hypothesis test. 

In the following section, we will use this theoretical result to specify the energy layer of our CERN.

  \begin{figure}[t!]
   \centering
   \begin{subfigure}{1.0\linewidth}
      \centering
      \includegraphics[width = 1.0\linewidth]{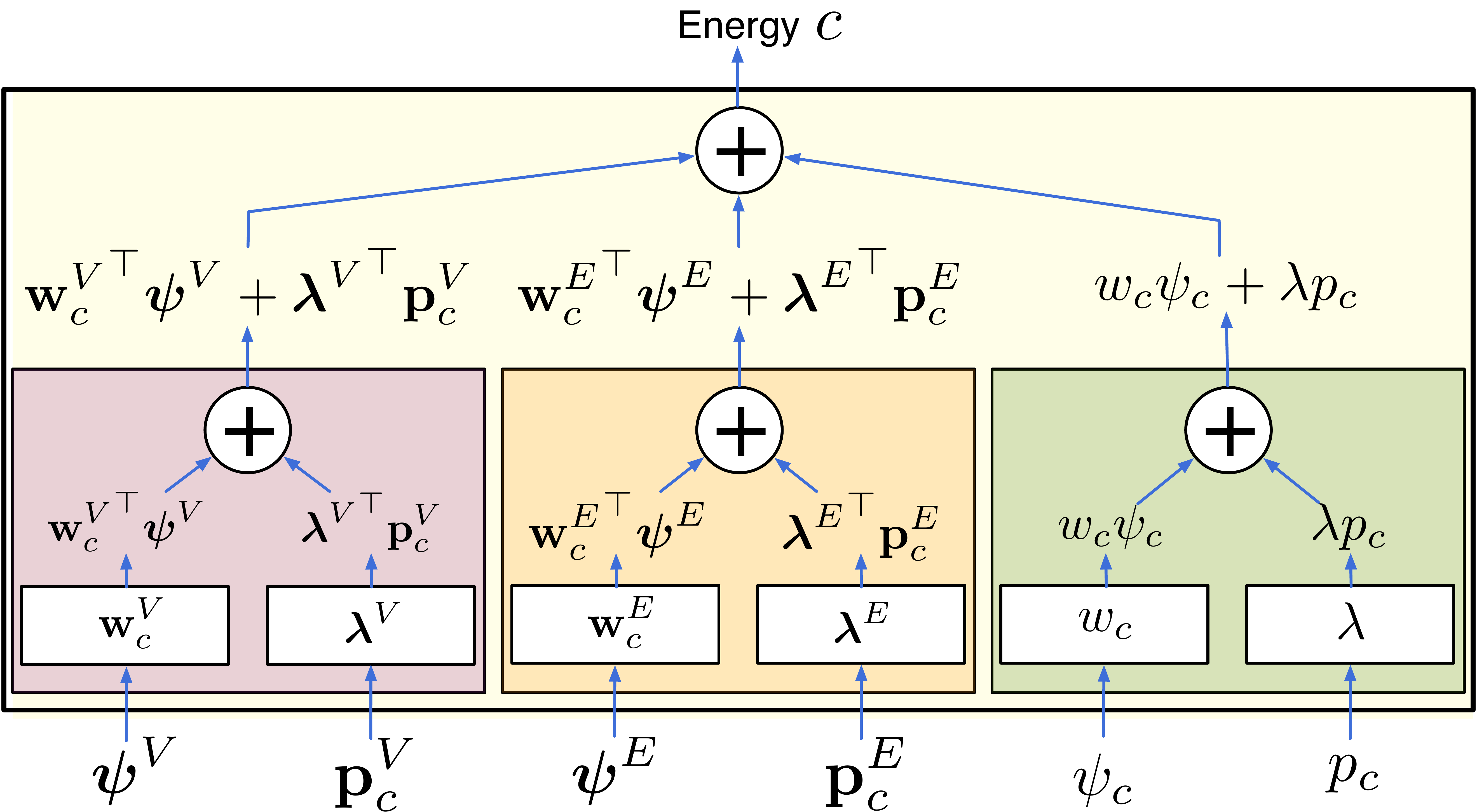}
	\caption{The unit for computing the regularized energy of category $c$, given by (\ref{eq:energy_c}).}
	\label{fig:energy_network_a}
	\end{subfigure}

   \begin{subfigure}{1.0\linewidth}
      \centering
      \includegraphics[width = 1.0\linewidth]{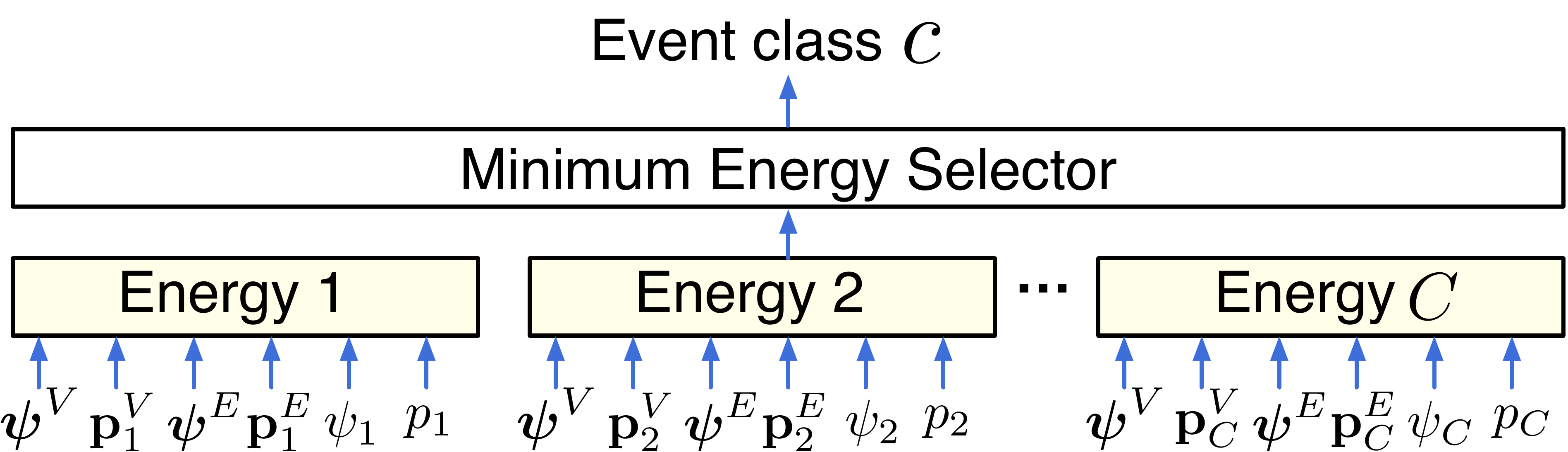}
	\caption{Diagram of all units in the energy layer.}
	\label{fig:energy_network_b}
   \end{subfigure}
      \caption{The EL takes the softmax outputs of all LSTMs along with estimated p-values as input, and outputs a solution that jointly minimizes the energy and maximizes a p-value of the Fisher's combined hypothesis test.}
      \label{fig:energy_network}
   \end{figure}

\section{The Energy Layer of CERN}
\label{sec:EL}
We extend the deep architecture of LSTMs with an additional energy layer (EL) aimed at jointly minimizing the energy, given by (\ref{eq:energy0_2}), and maximizing a p-value of the Fisher's combined hypothesis test, given by (\ref{eq:Fisher}).  For CERN-2, this optimization problem can be expressed as %
\begin{equation}
\begin{array}{lcl}
& \displaystyle \min_{c, Y} & \displaystyle \mathcal{E}(G)\\
 &\text{s.t.} & -\sum_{i\in V^\prime} \log p^V_i(c,y_i) \leq \tau^V,\\
 && -\sum_{(i,j) \in E^\prime} \log p^E_{ij}(c, y_{ij}) \leq \tau^E, \\
 && -\log p(c) < \tau^c,
 \end{array}
 \label{eq:optimization}
\end{equation}
where $\tau^V$, $\tau^E$, and $\tau^c$ are parameters that impose lower-bound constraints on the p-values. Recall that according to the Fisher's theory on a combined hypothesis test, decreasing the constraint parameters $\tau^V$, $\tau^E$, and $\tau^c$ will enforce higher p-values of the solution.

From (\ref{eq:energy0_2}) and (\ref{eq:optimization}), we derive the following Lagrangian, also referred to as  regularized energy $\tilde{\mathcal{E}}(X,Y, c)$, which can then be readily implemented as the EL:
\begin{equation}
\setlength{\arraycolsep}{0pt}
\begin{array}{l}
 \tilde{\mathcal{E}}(X,Y, c) =
 \displaystyle \sum_{i 
\in V}  w^V_{c,y_i}\psi^V(x_i,y_i){-} \lambda^V \sum_{i 
\in V}  \log p^V_i(c,y_i) \\ \displaystyle
+  \sum_{(i, j) \in E}w^E_{c,y_{ij}}  \psi^E(x_{ij},y_{ij}){-}  \lambda^E \sum_{(i, j) \in E}\log p^E_{ij}(c,y_{ij})\\
\displaystyle
+ w_c\psi(x, c) - \lambda\log p(c),
\end{array}
\label{eq:energy2}
\end{equation}
Note that for CERN-1, we drop the last two terms in (\ref{eq:energy2}), $w_c\psi_c$ and $\lambda\log p(c)$.
 $\tilde{\mathcal{E}}(X,Y, c)$ can be expressed in a more compact form as%
\begin{equation}
\setlength{\arraycolsep}{2pt}
\begin{array}{lcl}
 \tilde{\mathcal{E}}(X,Y, c) &=& {\ob_c^V}^\top\psib^V - {\lb^V}^\top \log\pb_c^V \\
 &&+ {\ob_c^E}^\top\psib^E - {\lb_c^E}^\top \log\pb_c^E \\
 &&+w_c\psi_c - \lambda\log p_c,
\end{array}
\label{eq:energy_c}
\end{equation}
where all parameters, potentials, and p-values are grouped into corresponding vectors. For brevity, we defer the specification of these vectors to the appendix.  

Fig.~\ref{fig:energy_network_a} shows a unit in the EL which computes (\ref{eq:energy_c}). After stacking these units, as shown in Fig.~\ref{fig:energy_network_b}, we select the solution $\hat{G}$ with the minimum $\tilde{\mathcal{E}}(\hat{G})$.

In the following, we explain our energy-based end-to-end training of the EL.

\section{Learning Regularized By Confidence}\label{sec:learning}

Following \cite{LeCun2006, Belanger2016}, we use an energy-based loss for a training instance $X^i$ and its ground truth labels $(Y^i, c^i)$ to learn parameters of the EL, i.e., the regularized energy, specified in (\ref{eq:energy2}):
\begin{equation}
\begin{array}{l}
L(X^i,Y^i, c^i)\\= \max\left(0, \tilde{\mathcal{E}}(X^i, Y^i, c^i) - \tilde{\mathcal{E}}(X^i, \bar{Y}, \bar{c}) +  \mathds{1}(c^{i} \neq \bar{c})\right),
\end{array}
\label{eq:loss}
\end{equation}
where $\bar{Y}, \bar{c} = \argmin_{Y, c\neq c^i}\tilde{\mathcal{E}}(X^i,Y, c) -  \mathds{1}(c^{i} \neq c)$ is the most violated case. Alternatively, this loss can be replaced by other energy-based loss functions also considered in \cite{LeCun2006}. Here we treat $Y$ as latent variables for simplicity and thus only consider accuracy of $c$. However, one can include a comparison between $Y$ and its corresponding ground truth label $Y^i$ into the loss function. It is usually difficult to find the most violated case. However, as \cite{LeCun2005} points out, the inference of the most violated case does not require a global minimum solution since the normalization term is not modeled in our energy-based model, so we can simply set $\bar{Y}$ to be the output of the node and edge LSTMs.  

In practice, one can first train a network using common losses such as cross-entropy to learn the representation excluding the EL, namely from the input layer to softmax layers. Then the p-value of a training instance can be computed by removing itself from the training sets $V_0$ and $E_0$. Finally we train the weights in (\ref{eq:energy2}) by minimizing the loss.

\section{Results}\label{sec:results}

{\bf Implementation details}. We stack the node LSTMs and edge LSTMs on top of a VGG-16 model \cite{Simonyan2014} without the FC-1000 layer. The VGG-16 is pre-trained on ImageNet \cite{Deng2009}, and fine-tuned with LSTMs jointly. We train the top layer of CERN by fixing the weights of the CNNs and the bottom layer LSTMs. The batch size for the joint training of the bottom LSTMs and VGG-16 is 6. The training converges within 20000 iterations. The event LSTM and the EL are trained using 10000 iterations with a batch size of 2000. For the mini-batch gradient descent, we use RMSprop \cite{Tieleman2012} with a learning rate ranging from $0.000001$ to $0.001$.  We use Keras \cite{chollet2015keras} with Theano \cite{Theano2016} as the backend to implement CERN, and run training and testing with a single NVIDIA Titan X (Pascal) GPU. For a fair comparison with \cite{Ibrahim2016}, we use the same tracker and its implementation as in \cite{Ibrahim2016}. Specifically, we use the tracker of \cite{Danelljan2014} from the Dlib library \cite{King2009}. The cropped image sequences of persons and pairs of persons are used as the inputs to node LSTMs and edge LSTMs, respectively.

We compare our approach with the state-of-the-art methods \cite{Hajimirsadeghi2015, Ibrahim2016}. In addition, we evaluate the following reasonable baselines.

{\bf Baselines:}
\begin{itemize}[itemsep=-5pt,topsep=2pt, partopsep=1pt]
\item 2-layer LSTMs (B1).  We test a network of 2-layer LSTMs similar to \cite{Ibrahim2016}. All other baselines below and our full models use B1 to compute their potentials and p-values. B1 does not have the energy layer, but only a feed-forward network. The event class is predicted by the softmax output of the event LSTM.
\item CERN-1 w/o p-values (B2). This baseline represents the CERN-1 network with the EL, however, the p-values are not computed and not used for regularizing energy minimization. Hence, the event class prediction of B2 comes from the standard energy minimization.
\item CERN-2 w/o p-values (B3). Similar to B2, in this B3, we do not estimate and do not use the p-values in the EL of CERN-2.
\end{itemize}

{\bf Datasets}. We evaluate our method in two domains: collective activities and sport events using the Collective Activity dataset \cite{Choi2009} and the Volleyball dataset \cite{Ibrahim2016} respectively.

\subsection{Collective Activity Dataset}

The Collective Activity dataset consists of 44 videos, annotated with 5 activity categories (\textit{crossing}, \textit{walking}, \textit{waiting}, \textit{talking}, and \text{queueing}), 6 individual action labels (\textit{NA}, \textit{crossing}, \textit{walking}, \textit{waiting}, \textit{talking}, and \text{queueing}), and 8 pairwise interaction labels (\textit{NA}, \textit{approaching}, \textit{leaving}, \textit{passing-by}, \textit{facing-each-other}, \textit{walking-side-by-side}, \textit{standing-in-a-row}, \textit{standing-side-by-side}). The interaction labels are provided by the extended annotation in \cite{Choi2014}. 

For this dataset, we first train the node LSTMs and edge LSTMs with 10 time steps and 3000 nodes. Then, we concatenate the outputs of these two types of LSTMs at the bottom layer of CERN, along with their VGG-16 features, and pass the concatenation to the bidirectional event LSTM with 500 nodes and 10 time steps at the top layer of CERN. The concatenation is passed through a max pooling layer and a fully-connected layer with a output dimension of 4500. 

For comparison with \cite{Hajimirsadeghi2015, Ibrahim2016} and baselines B1-B3, we use the following performance metrics: (i) multi-class classification accuracy (MCA), and (ii) mean per-class accuracy (MPCA). Our split of training and testing sets is the same as in \cite{Hajimirsadeghi2015, Ibrahim2016}. Tab.~\ref{table:collective} summarizes the performance of all methods on recognizing group activities. Note that in Tab.~\ref{table:collective}  only \cite{Hajimirsadeghi2015} does not use deep neural nets. As can be seen, our energy layer significantly boosts the accuracy, outperforming the state-of-the-art by a large margin. Even when we only have the bottom layer of LSTMs, CERN-1 still outperforms the 2-layer LSTMs in \cite{Ibrahim2016} thanks to the EL. Without the EL, the baseline B1 yields lower accuracy than \cite{Ibrahim2016} even with additional LSTMs for the interactions.

Our accuracies of recognizing individual actions and interactions on the Collective Activity dataset are 72.7\% and 59.9\%, using the node LSTMs and edge LSTMs respectively. Note that B1, CERN-1 and CERN-2 share the same node and edge LSTMs. 

\begin{figure}[t!]
      \centering
      \includegraphics[clip,trim={50 5 45 5}, width = 0.9\linewidth]{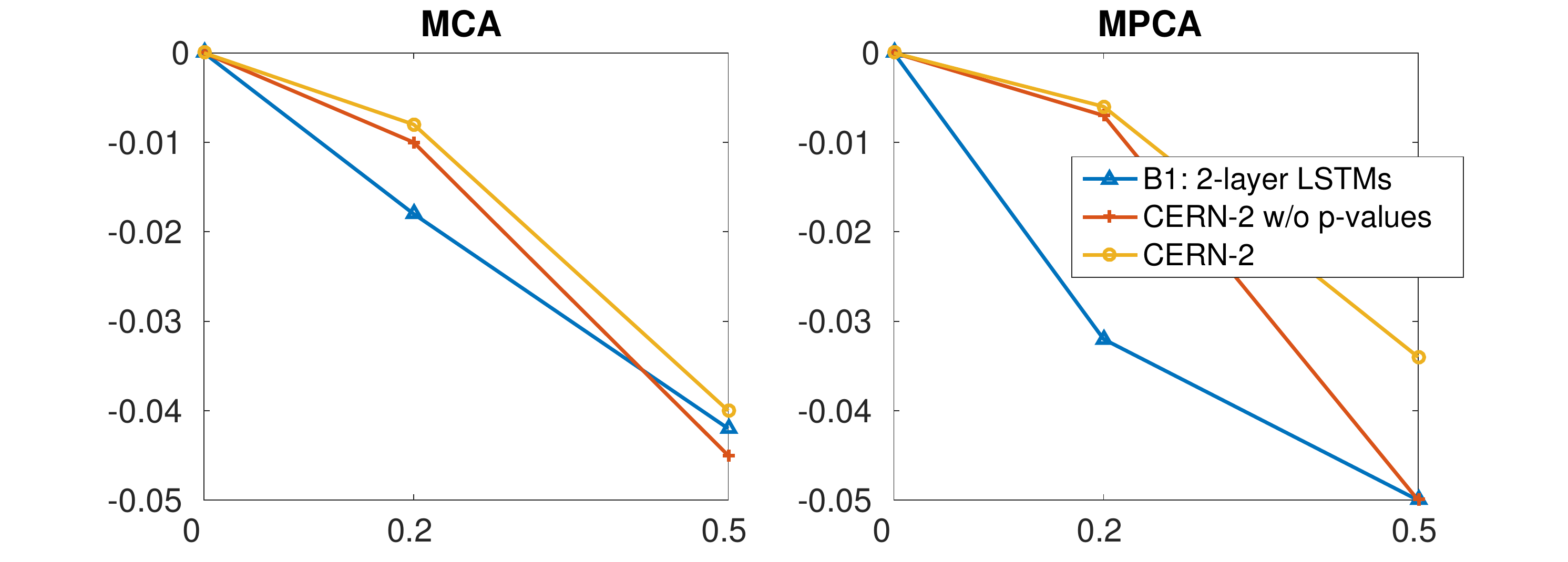}
      \caption{Performance decrease of group activity recognition for a varying percentage of corruption of human trajectories in the Collective Activity dataset. We compare 2-layer LSTMs (B1), CERN-2 w/o p-values (B3) and CERN-2 using the same corrupted trajectories as input.}
      \label{fig:stability_collective}
   \end{figure}

For evaluating numerical stability of predicting group activity classes by CERN-2, we corrupt all human trajectories in the testing data, and control the amount of corruption with the corruption probability. For instance, for the corruption probability of 0.5, we corrupt one bounding box of a person in every video frame with a 0.5 chance. When the bounding box is selected, we randomly shift it with a horizontal and a vertical displacement ranging from 20\% to 80\% of the original bounding box's width and height respectively. As Fig.~\ref{fig:stability_collective} shows, CERN-2 consistently experiences a lower degradation in performance compared to the baselines without p-values. This indicates that incorporating the p-values into the energy model indeed benefits the inference stability. Such benefit becomes more significant as the amount of corruption in input data increases.

\begin{figure*}[t!]
      \centering
      \includegraphics[clip,trim={0 0 0 0}, width = 0.7\linewidth]{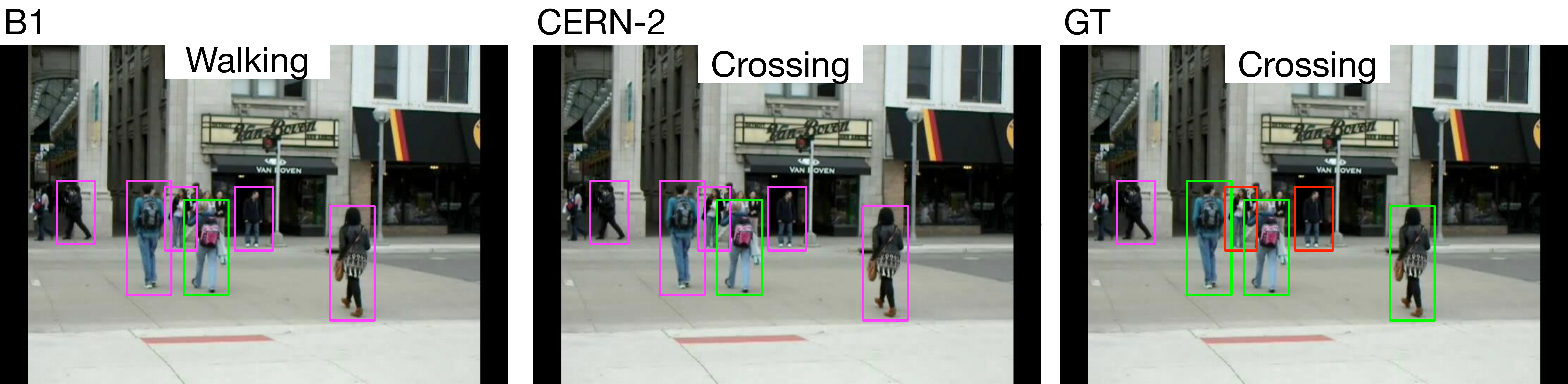}
      \caption{The qualitative results on the Collective Activity dataset. From left to right, we show the inference results from B1, CERN-2 and the ground truth (GT) labels respectively. The colors of the bounding boxes indicate the individual action labels (green: \textit{crossing}, red: \textit{waiting}, magenta: \textit{walking}). The interaction labels are not shown here for simplicity.}
      \label{fig:qualitative_collective}
   \end{figure*}

Fig.~\ref{fig:qualitative_collective} shows an example of the \textit{crossing} activity. As can be seen, although B1 and CERN-2 share the same individual action labels, where a majority of the people are assigned incorrect action labels, CERN-2 can still correctly recognize the activity.

\begin{table}
\centering
\begin{tabular}{|l|c|c|}
\hline
Method & MCA & MPCA\\ \hhline{===}
Cardinality kernel \cite{Hajimirsadeghi2015} 	&	 83.4 	&	81.9\\ \hhline{===}
2-layer LSTMs \cite{Ibrahim2016}			&	 81.5 	&	80.9\\ \hline
B1: 2-layer LSTMs 						&	 79.7		&	80.3\\ \hline
B2: CERN-1 w/o p-values					&	 83.8		& 	84.3\\ \hline
B3: CERN-2 w/o p-values					&	 83.8		& 	83.7\\ \hline
CERN-1								&	 84.8		&	85.5\\ \hline
CERN-2								& {\bf 87.2}	&{\bf 88.3}\\ \hline
\end{tabular}
\caption{Comparison of different methods for group activity recognition on the Collective Activity dataset.}
\label{table:collective}
\end{table}

\begin{table} 
\centering
\begin{tabular}{|l|c|c|}
\hline
Method & MCA & MPCA\\ \hhline{===}
2-layer LSTMs \cite{Ibrahim2016} (1 group)		&	 70.3 	&	65.9\\ \hline
B1: 2-layer LSTMs (1 group)					&	 71.3		&	69.5\\ \hline
B2: CERN-1 w/o p-values	(1 group)				&	 33.3		& 	34.3\\ \hline
B3: CERN-2 w/o p-values	(1 group)				&	 71.7		& 	69.8\\ \hline
CERN-1 (1 group)							&	 34.4		& 	34.9\\ \hline
CERN-2 (1 group)							& 	 73.5		&	72.2\\ \hhline{===}
2-layer LSTMs \cite{Ibrahim2016} (2 groups)		&	 81.9 	&	82.9\\ \hline
B1: 2-layer LSTMs (2 group)					&	 80.3		&	80.5\\ \hline
B3: CERN-2 w/o p-values	(2 groups)				&	 82.2		& 	82.3\\ \hline
CERN-2 (2 groups)							& 	 \bf{83.3}	&	\bf{83.6}\\ \hline
\end{tabular}
\caption{Comparison of different methods for group activity recognition on the Volleyball dataset. The first block is for the methods with 1 group and the second one is for those with 2 groups.}
\label{table:volleyball}
\end{table}

      \begin{figure}[t!]
      \centering
      \includegraphics[clip,trim={60 5 60 5}, width = 0.9\linewidth]{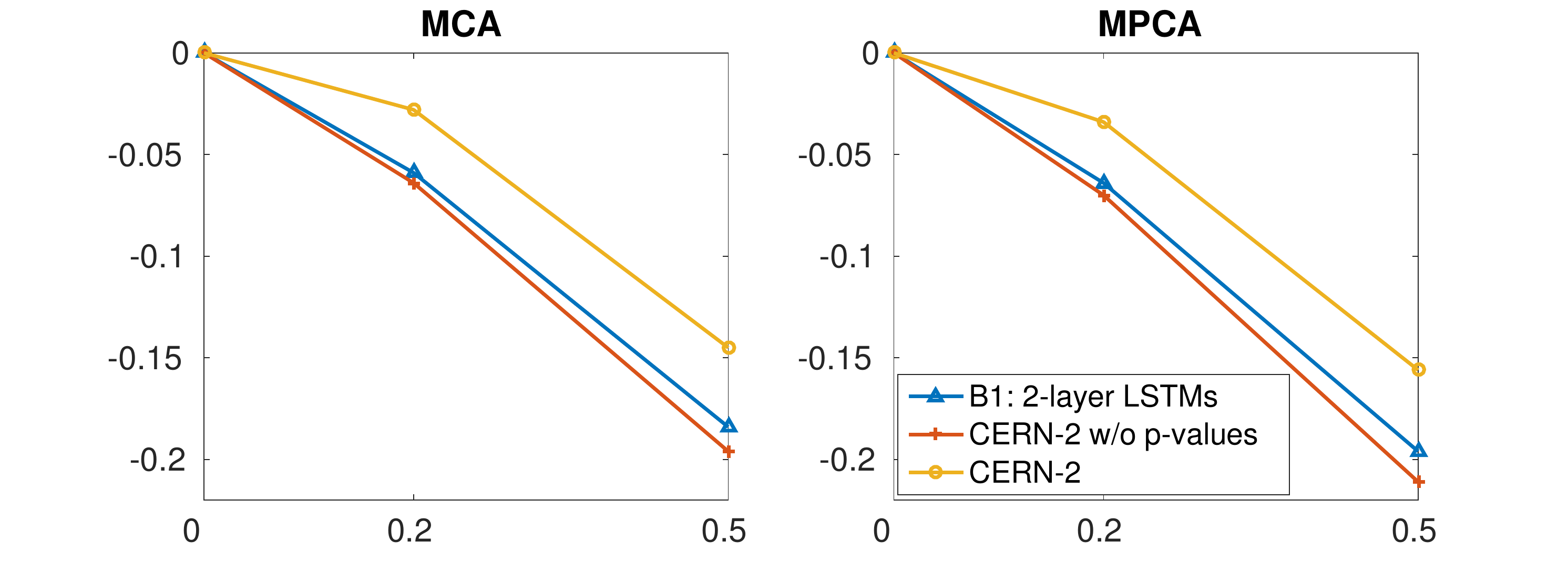}
      \caption{The decrease of group activity recognition accuracy over different input distortion percentages on the Volleyball dataset (all use the 2 groups style). CERN-2 is compared with 2-layer LSTMs (B1) and CERN-2 w/o p-values (B3).}
      \vspace{-10pt}
      \label{fig:stability_volleyball}
   \end{figure}

         \begin{figure*}[t!]
      \centering
      \includegraphics[clip,trim={50 80 100 0}, width = 0.95\linewidth]{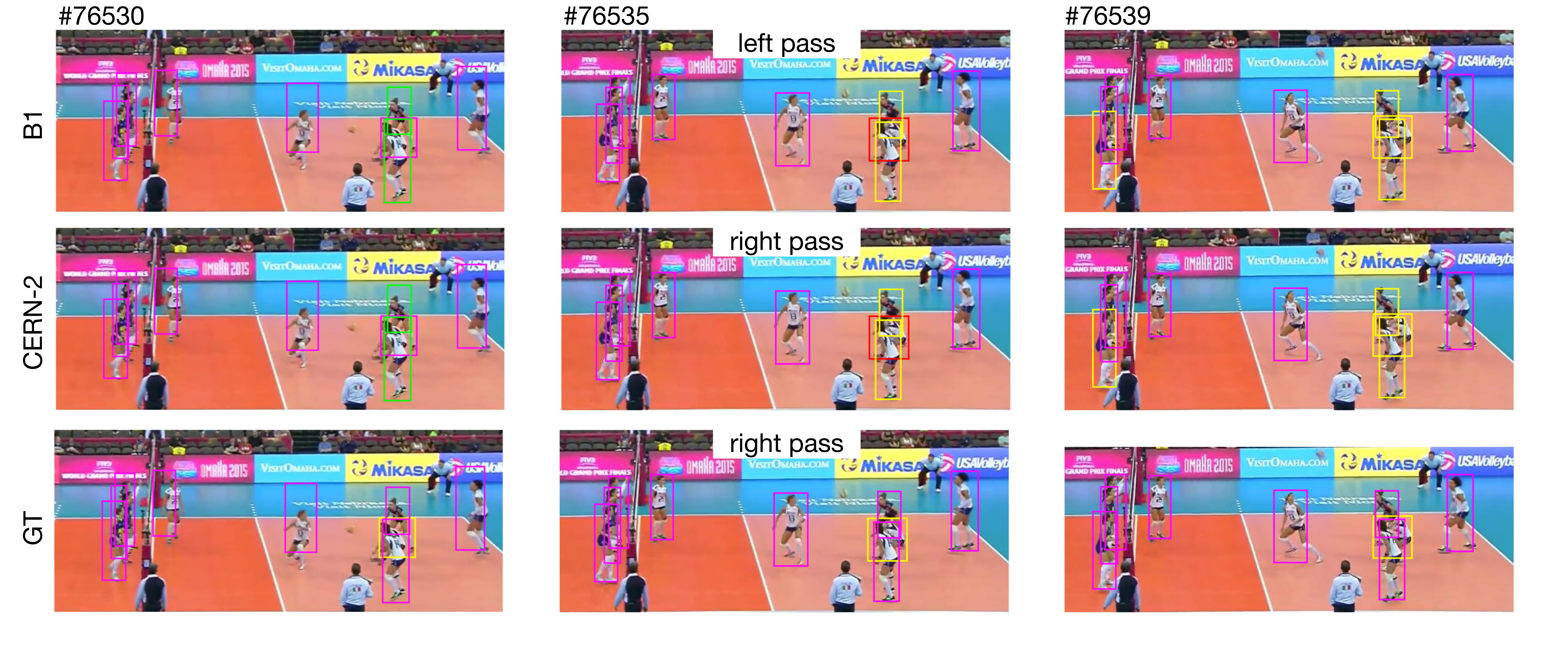}
      \caption{The qualitative results on the Volleyball dataset: results of B1 (top), results of CERN-2 (middle) and the ground truth (GT) labels (bottom). The colors of the bounding boxes indicate the individual action labels (green: \textit{waiting}, yellow: \textit{digging}, red: \textit{falling}, magenta: \textit{standing}), and the numbers are the frame IDs.}
      \label{fig:qualitative_volleyball}
   \end{figure*}

\subsection{Volleyball Dataset}

The Volleyball dataset consists of 55 videos with 4830 annotated frames. The actions labels are \textit{waiting}, \textit{setting}, \textit{digging}, \textit{failing}, \textit{spiking}, \textit{blocking}, \textit{jumping}, \textit{moving}, and \textit{standing}; and the group activity classes include \textit{right set}, \textit{right spike}, \textit{right pass}, \textit{right winpoint}, \textit{left winpoint}, \textit{left pass}, \textit{left spike}, and \textit{left set}. Interactions are not annotated in this dataset, so we do not recognize interactions and remove the edge LSTMs.

The node LSTMs have 3000 nodes and 10 time steps (including 5 preceding and 4 succeeding frames). The event LSTM in CERN-2 is a bidirectional LSTM with 1000 nodes and 10 time steps. In \cite{Ibrahim2016}, the max pooling has two types: 1) pooling over the output of all node LSTMs, or 2) dividing the players into two groups (the left team and the right team) first and pooling over each group separately. We test both types of max pooling for our approach to rule out the effect of pooling type in the comparison. CERN-1 does not have the pooling layer, thus is categorized as 1 group style. 

Recognition accuracy of individual actions is 69.1\% using node LSTMs, and the accuracies of recognizing group activities are summarized in Tab.~\ref{table:volleyball}. Cleary, the regularized energy minimization increases the accuracy compared to the conventional energy minimization (B2 and B3), and CERN-2 outperforms the state-of-the-art when using either of the pooling types. CERN-1 does not achieve accuracy that is comparable to that of CERN-2 on the Volleyball dataset. This is mainly because CERN-1 reasons the group activity based on individual actions, which may not provide sufficient information for recognizing complex group activities in sports videos. CERN-2 overcomes this problem by adding the event LSTM.

We also evaluate the stability of recognizing group activities by CERN-2 under corruption of input human trajectories. As Fig.~\ref{fig:stability_volleyball} indicates, the p-values in the EL indeed increase the inference reliability on the Volleyball dataset.

The qualitative results (2 groups) of a \textit{right pass} activity is depicted in Fig.~\ref{fig:qualitative_volleyball}, which demonstrates the advantage of the inference based on our regularized energy compared to the softmax output of the deep recurrent networks when the action predictions are not accurate.

\section{Conclusion}

We have addressed the problem of recognizing group activities, human interactions, and individual actions with a novel deep architecture, called Confidence-Energy Recurrent Network (CERN). CERN extends an existing two-level hierarchy of LSTMs by additionally incorporating a confidence measure and an energy-based model toward improving reliability and numerical stability of inference. Inference is formulated as a joint minimization of the energy and maximization of the confidence measure of predictions made by the LSTMs. This is realized through a new differentiable energy layer (EL) that computes the energy regularized by a p-value of the Fisher's combined statistical test. We have defined an energy-based loss in terms of the regularized energy for learning the EL  end-to-end. CERN has been evaluated on the Collective Activity dataset and Volleyball dataset. In comparison with previous approaches that predict group activities in a feed-forward manner using deep recurrent networks, CERN gives a superior performance, and also gives more numerically stable solutions under uncertainty. For collective activities, our simpler variant CERN-1 gives more accurate predictions than a strong baseline representing a two-level hierarchy of LSTMs with softmax outputs taken as predictions. Our variant CERN-2 increases complexity but yields better accuracy on challenging group activities which are not merely a sum of individual actions but a complex whole.

\section*{Acknowledgements}
This research was supported by grants DARPA MSEE project FA 8650-11-1-7149, ONR MURI project N00014-16-1-2007, and NSF IIS-1423305.

{\small
\bibliographystyle{ieee}
\bibliography{egbib}
}

\appendix
\section*{Appendix}

\section{Energy Function}

The regularized energy $\tilde{\mathcal{E}}(X,Y, c)$ can be reformulated in a compact form as%
\begin{equation}
\setlength{\arraycolsep}{2pt}
\begin{array}{lcl}
 \tilde{\mathcal{E}}(X,Y, c) &=& {\ob_c^V}^\top\psib^V - {\lb^V}^\top \log\pb_c^V \\
 &&+ {\ob_c^E}^\top\psib^E - {\lb_c^E}^\top \log\pb_c^E \\
 &&+w_c\psi_c - \lambda\log p_c,
\end{array}
\label{eq:energy_c}
\end{equation}
where weights of the EL are grouped into $\{w_c\}_{c =1,\cdots,C}$, $\lambda$, and the following parameter vectors:
\begin{equation}
\setlength{\arraycolsep}{0pt}
\begin{array}{ll}
\ob_c^V = \left[w_{c,1}^V, \cdots, w_{c,|\mathcal{Y}^V|}\right]^\top,&
\ob_c^E = \left[w_{c,1}^E, \cdots, w_{c,|\mathcal{Y}^E|}\right]^\top,\\
\lb^V = \left[\lambda^V, \cdots, \lambda^V\right]^\top,&
\lb^E = \left[\lambda^E, \cdots, \lambda^E\right]^\top,
\end{array}
\label{eq:parameters}
\end{equation}
and the input to the EL is specified in terms of the LSTM softmax outputs and p-values:
\begin{equation}
\setlength{\arraycolsep}{0pt}
\begin{array}{l}
\displaystyle
\psib^V = \Big[\sum_{i : y_i = 1} \psi^V(x_i, y_i), \cdots, \sum_{i:y_i = |\mathcal{Y}^V|} \psi^V(x_i, y_i)\Big]^\top,\\
\displaystyle
\psib^E=  \Big[\sum_{\substack{(i,j):\\y_{ij} = 1}} \psi^E(x_{ij}, y_{ij}), \cdots, \sum_{\substack{(i,j):\\y_{ij} = |\mathcal{Y}^E|}} \psi^E(x_{ij}, y_{ij})\Big]^\top,\\\displaystyle
\psi_c = \psi(x, c),\\\displaystyle
\pb_c^V = \Big[\sum_{i : y_i = 1} p_i^V(c, y_i), \cdots, \sum_{i: y_i = |\mathcal{Y}^V|} p_i^V(c, y_i)\Big]^\top,\\\displaystyle
\pb_c^E = \Big[\sum_{(i,j) : y_{ij} = 1} p_{ij}^V(c, y_i), \cdots, \sum_{(i,j): y_{ij} = |\mathcal{Y}^E|} p_{ij}^V(c, y_i)\Big]^\top,\\
p_c = p(c).
\end{array}
\end{equation}

\end{document}